\documentclass[sigconf]{acmart}
\usepackage{dsfont}
\usepackage{lineno,hyperref}
\usepackage[ruled]{algorithm2e}  
\usepackage{algpseudocode}
\graphicspath{{fig/}}
\usepackage{colortbl}
\usepackage{multirow}

\def\eg{\emph{e.g.}}

\def\al{\emph{et al. }}

\AtBeginDocument{
	\providecommand\BibTeX{{
			\normalfont B\kern-0.5em{\scshape i\kern-0.25em b}\kern-0.8em\TeX}}}

\copyrightyear{2020}
\acmYear{2020}
\setcopyright{acmcopyright}
\acmConference[MM '20]{Proceedings of the 28th ACM International Conference on Multimedia}{October 12--16, 2020}{Seattle, WA, USA}
\acmBooktitle{Proceedings of the 28th ACM International Conference on Multimedia (MM '20), October 12--16, 2020, Seattle, WA, USA}
\acmPrice{15.00}
\acmDOI{10.1145/3394171.3414044}
\acmISBN{978-1-4503-7988-5/20/10}

\SetKwInput{KwInput}{Input}        
\SetKwInput{KwOutput}{Output}

\begin{document}

\fancyhead{}

\title{Data-driven Meta-set Based Fine-Grained Visual Classification}

\author{Chuanyi Zhang}
\affiliation{
\institution{Nanjing University of Science and Technology, Nanjing, China}
}
\email{zhangchuanyi@njust.edu.cn}

\author{Yazhou Yao}
\authornote{Corresponding author}
\affiliation{
\institution{Nanjing University of Science and Technology, Nanjing, China}}
\email{yazhou.yao@njust.edu.cn}

\author{Xiangbo Shu}
\affiliation{
\institution{Nanjing University of Science and Technology, Nanjing, China}}
\email{shuxb@njust.edu.cn}

\author{Zechao Li}
\affiliation{
\institution{Nanjing University of Science and Technology, Nanjing, China}}
\email{zechao.li@njust.edu.cn}

\author{Zhenmin Tang}
\affiliation{
\institution{Nanjing University of Science and Technology, Nanjing, China}}
\email{tzm.cs@njust.edu.cn}

\author{Qi Wu}
\affiliation{
	\institution{University of Adelaide, Adelaide, Australia}
}
\email{qi.wu01@adelaide.edu.au}

\begin{abstract}
		
Constructing fine-grained image datasets typically requires domain-specific expert knowledge, which is not always available for crowd-sourcing platform annotators. Accordingly, learning directly from web images becomes an alternative method for fine-grained visual recognition. However, label noise in the web training set can severely degrade the model performance. To this end, we propose a data-driven meta-set based approach to deal with noisy web images for fine-grained recognition. Specifically, guided by a small amount of clean meta-set, we train a selection net in a meta-learning manner to distinguish in- and out-of-distribution noisy images. To further boost the robustness of model, we also learn a labeling net to correct the labels of in-distribution noisy data. In this way, our proposed method can alleviate the harmful effects caused by out-of-distribution noise and properly exploit the in-distribution noisy samples for training. Extensive experiments on three commonly used fine-grained datasets demonstrate that our approach is much superior to state-of-the-art noise-robust methods. The data and source code of this work have been made available at: {\url{https://github.com/NUST-Machine-Intelligence-Laboratory/dmbfgvr}}.
		
\end{abstract}

\ccsdesc[500]{Computing methodologies~Computer vision}
	
\keywords{label noise; fine-grained; in-distribution; out-of-distribution}
	
\maketitle
	
\section{Introduction}
	
Deep Neural Networks (DNNs) have achieved impressive results on many computer vision tasks due to the availability of large-scale image datasets \cite{yao2018tip,yao2018extracting,xie2019attentive,luo2019segeqa,xie2020eccv,2018extracting,lu2020hsi,chen2020classification,tang2017personalized,shu2019hierarchical}. However, fine-grained visual classification (FGVC) remains challenging. Training DNNs for FGVC tends to require expert-annotated labels, possibly with additional annotations in the form of parts, attributes, or relationships. The high cost of manual annotation limits the FGVC dataset scale and constrains the performance and scalability of the model. To reduce the cost of manual labeling, a growing number of works focused on the semi-supervised paradigm \cite{xu2015augmenting,niu2018webly,cui2016fine} and utilized web images as data augmentation.
	
Web images have distinct advantages over manual labeled ones: rich and free. For arbitrary categories, the potential training data can be easily obtained from publicly-available sources like Google Image. Leveraging web data can easily build large-scale datasets at nearly no manual cost \cite{yao2017exploiting,yao2019towards,yao2016domain,yao2017new,yao2016automatic}. Unfortunately, web data inevitably contains label noise which is a huge obstacle for training robust deep FGVC models. The label noise in web data for fine-grained recognition can be roughly divided into two sets: in-distribution and out-of-distribution. Specifically, the in-distribution noisy images have their true labels in the dataset, while the true labels for out-of-distribution noisy samples are outside of the dataset. Since DNNs have a high capacity to fit noisy data \cite{arpit2017closer,zhang2016understanding}, directly utilizing noisy web images to train fine-grained recognition models usually results in poor performance.
	
A simple yet effective approach to deal with label noise is to perform samples selection that separates clean instances from noise. The representative works are Decoupling \cite{malach2017decoupling} and Co-teaching \cite{han2018co}. These works drop samples with a high probability of being incorrectly labeled to reduce the harmful influence of noise. Nevertheless, they can't exploit the in-distribution noisy instances for representation learning and have the risk to discard some clean images. To make full use of the noisy dataset, some works concentrated on loss correction that revises the corrupted labels. For example, \cite{goldberger2016training} added an additional softmax layer to estimate the label noise transition matrix. Unfortunately, since the label noise in practice is diverse and non-stationary, the exact recovery of the noise transition matrix is difficult and remains a challenging problem. \cite{song2019selfie} selectively refurbished unclean samples which have consistent label predictions. However, the network has the risk to produce wrong labels. Moreover, these loss correction methods are unable to tackle out-of-distribution noisy images, whose true labels are outside the set of training labels.
	
In this work, we propose a hybrid approach that leverages the advantages of both "sample selection" and "loss correction" to learn from noisy web images for fine-grained task. We make an assumption that the model can access a small set of clean meta images during training. Our key idea is to discard out-of-distribution noise as well as to relabel in-distribution noisy samples with the help of small clean meta-data. Specifically, we train a selection net $ S_{net} $ to learn the similarities between noisy web images and clean meta set. It produces the probability that a sample is in-distribution and we drop these images which have a low in-distribution probabilities. Simultaneously, we train a labeling net $ L_{net} $ to generate pseudo labels for the remaining in-distribution noisy images. In this way, our proposed approach can learn from the noisy web images with the guidance of clean meta data. It can properly utilize in-distribution noisy images for training and alleviate the harmful effects caused by out-of-distribution noise. 

In summary, this paper makes the following three-fold contributions:	
(1) We propose a data-driven meta-set based approach that combines the "sample selection" and "loss correction" methods by overcoming their drawbacks. Compared with sample selection approaches, our approach can relabel and exploit the in-distribution noisy samples for boosting training. Compared with loss correction methods, our method is less likely to suffer from the correction error and can deal with out-of-distribution noisy samples.	
(2) We explained the effectiveness of our proposed approach from the perspective of mathematical theory. Our proposed method has a better interpretability.	
(3) Extensive experiments and ablation studies demonstrate that our approach outperforms state-of-the-art methods.
	
\section{Related Works}
	
\textbf{Fine-grained Visual Classification:} 
The task of fine-grained visual classification is to distinguish objects at the subordinate level. Since differences between subcategories tend to lie in discriminative parts, the early works train the network to learn discriminative features by utilizing strong annotation like bounding boxes or part annotations \cite{huang2016part,yao2016coarse,lam2017fine,wei2018mask,zhang2014part,xie2015hyper}. Despite satisfying performance, these strongly supervised methods require heavy human annotation. To avoid heavy manual labeling work, there have been proposed a number of weakly supervised methods, which only need image-level labels \cite{fu2017look,lin2015bilinear,zheng2017learning,wang2018learning,ge2019weakly,zheng2019looking,chen2019destruction,korsch2019classification,zhang2016picking,he2017fine,peng2017object,zhang2016weakly,branson2014bird}. However, the label annotation still requires expert knowledge. This drawback limits the dataset scale and therefore constrains the model performance and scalability. To further improve the performance, some semi-supervised methods manage to leverage easily accessible web data \cite{niu2018webly,cui2016fine,xu2016webly,niu2015visual,xiao2015learning,krause2016unreasonable,van2015building} for FGVC. However, these methods mainly utilize web images as data augmentation and still require a large number of well-labeled images. Different from them, our approach merely requires a small set of clean images to guide noise identification and label correction, and the model is mainly trained from web images.
	
\textbf{Learning from Web Images:}	
Training fine-grained recognition models with web images usually results in poor performance due to the presence of label noise and data bias. Numerous studies have been performed to address the problem of learning from noisy web images \cite{yao2020exploiting,2018discovering,2019dynamically,zhang2020web}. In this paper, we categorize them into two groups: sample selection based methods and loss correction based methods. Sample selection methods are a straightforward way to tackle label noise by discarding the noisy ones. The representative sample selection methods contain Decoupling \cite{malach2017decoupling} and Co-teaching \cite{han2018co}. They both trained two peer networks simultaneously. Specifically, Decoupling chooses samples that have different predictions from two networks as useful ones. Co-teaching lets each network selects small-loss samples as clean ones for its peer network. These methods have achieved remarkable performance on noisy data by ignoring all unclean samples. However, they fail to utilize potentially useful in-distribution noisy images and have the risk to eliminate clean samples. Loss correction approaches aim to correct the misguidance caused by label noise. The correction methods are diverse, including assigning a weight to the current prediction \cite{reed2014training}, estimating label noise transition \cite{goldberger2016training}, and relabeling samples with the network prediction \cite{song2019selfie}. Although these methods have shown significant performance on the manual noisy dataset, they are unable to tackle out-of-distribution samples in noisy web data. Our method is a combination of sample selection and loss correction approach by achieving their advantages and overcoming their drawbacks. It can robustly learn from noisy web images in the real-world dataset.

\textbf{Data-driven Approaches:}	
Considering that learning merely from noisy web images is challenging, some works make the assumption that during training the model has access to a small set of clean labels \cite{li2020weakly,li2018deep}. For example, \cite{hendrycks2018using} leveraged a small set of clean data to estimate the noise transition matrix. \cite{li2017learning}  distilled the knowledge learned from the clean dataset to facilitate learning a better model from the entire noisy dataset. \cite{ren2018learning} and \cite{shu2019meta} utilized a meta-learning algorithm to reweight training examples. Specifically, \cite{ren2018learning} dynamically learned weights based on samples' gradient directions, while \cite{shu2019meta} learned an explicit weighting function. With access to a small set of clean samples, these data-driven methods achieve remarkable performance and robustness. Inspired by them, our approach leverages a small number of clean images to guide sample selection and label correction.

\begin{figure*}[t]
	\centering 
	\includegraphics[width=0.99\textwidth]{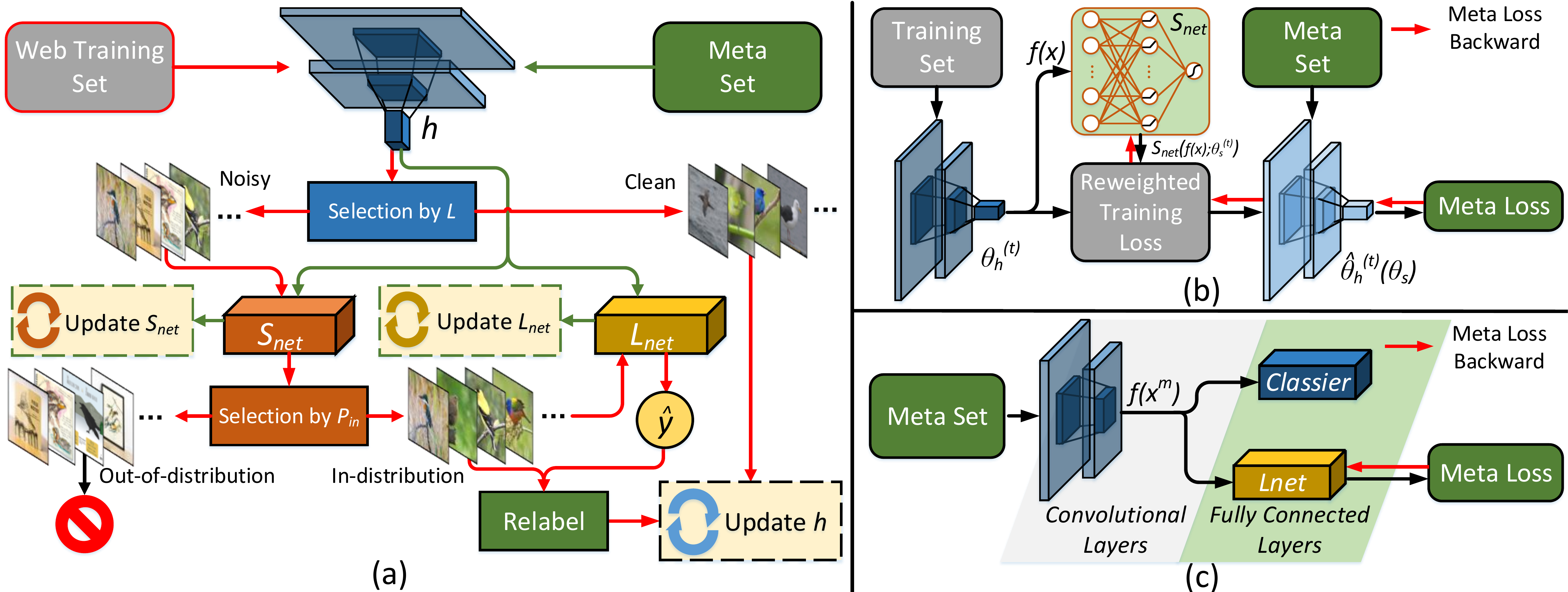} 
	\caption{The architecture of our proposed data-driven meta-learning based approach (a), training mechanism of selection net $ S_{net} $ (b), and training mechanism of labeling net $ L_{net} $ (c). For each mini-batch from the noisy web training set, we first split it by training loss $ \mathcal{L} $. Small $ \mathcal{L} $ images are regarded as clean data and directly utilized for updating the networks. Samples with large $ \mathcal{L} $ are entered into $ S_{net} $ to compute their probabilities of being in-distribution $ P_{in} $. Then samples with small $ P_{in} $ are identified as out-of-distribution noisy images and discarded. Those with large $ P_{in} $ are regarded as in-distribution noisy images. They are relabeled by $ L_{net} $ and leveraged for training with their pseudo labels $ \hat{y} $. Our $ S_{net} $ and $ L_{net} $ both take the image features as input and are updated using the meta set.}
	\label{fig1}
	\vspace{-0.3cm}
\end{figure*}
	
\section{The Proposed Approach}
	
\subsection{Overview}
	
The architecture of our proposed framework is presented in Fig.~\ref{fig1} (a). Let $(x_{i}, y_{i})$ be the pair of sample $ x_{i} $ and its label $ y_{i} $, $ \mathcal{D} = \{(x_{i}, y_{i})|1 \leq i \leq N\} $ be the noisy web training set. We assume that there is a small clean and unbiased meta set $ \mathcal{D}^{m} = \{(x_{i}^{m}, y_{i}^{m})|1 \leq i \leq M\} $, where $ M \ll N $. In addition, we also assume that the training set contains the meta set. Our goal is to train the classifier network $ h $ with the parameter $ \theta_{h} $ on the noisy web training set.
	
Since deep neural networks have the ability to filter out noisy instances using their loss values at the early training stage \cite{arpit2017closer,zhang2016understanding}, we simply utilize the entire training set to train the network in initial epochs. When a mini-batch $ \overline{\mathcal{D}} $ is formed from $ \mathcal{D} $, $ \theta_{h} $ is updated through:
\begin{equation}
	\theta_{h}^{(t+1)} = \theta_{h}^{(t)} - \frac{\alpha}{|\overline{\mathcal{D}}|} \nabla \sum_{x \in \overline{\mathcal{D}}} \mathcal{L}(x,y;\theta_{h}^{(t)}), T \leq T_{s},
	\label{eq1}
\end{equation}
where $ \mathcal{L} $, $ \alpha $, $ T $ and $ T_{s} $ denote the loss function, learning rate, epoch number and initial epoch number, respectively. 
	
After the initial epochs, we start to adopt our proposed algorithm to tackle noise. We first adopt the widely used loss-based separation method \cite{malach2017decoupling} that selects $ (1-\tau) \times 100\% $ of low-loss instances as clean samples. The mini-batch is divided into a clean set $ \overline{\mathcal{D}}^{c} $ and a noisy set $ \overline{\mathcal{D}}^{n} $, which can be obtained by solving the following problem:
\begin{equation}
	\overline{\mathcal{D}}^{c} = \mathop{\arg\min}_{\hat{\mathcal{D}}:|\hat{\mathcal{D}}| \geq (1-\tau)|\overline{\mathcal{D}}|} \sum\nolimits_{x \in \hat{\mathcal{D}}} \mathcal{L}(x,y;\theta_{h}),
	\label{eq2}
\end{equation}
\begin{equation}
	\overline{\mathcal{D}}^{n} = \overline{\mathcal{D}} - \overline{\mathcal{D}}^{c},
	\label{eq3}
\end{equation}
where $ \tau $ is the drop rate. Then $ \overline{\mathcal{D}}^{c} $ is directly leveraged for training, while images in $ \overline{\mathcal{D}}^{n} $ are entered into the selection net $ S_{net} $ to calculate their probability to be in-distribution $ P_{in} $ via:
\begin{equation}
	P_{in}(x) = S_{net}(f(x);\theta_{s}), x \in \overline{\mathcal{D}}^{n},
	\label{eq4}
\end{equation}
where $ f(x) $ denotes the features of sample $ x $ and $ \theta_{s} $ is the parameter of $ S_{net} $. Furthermore, we select $ r \times 100\% $ samples that have high $ P_{in} $ as in-distribution noisy images and discard others. The in-distribution set $ \overline{\mathcal{D}}^{in} $ can be obtained by:
\begin{equation}
	\overline{\mathcal{D}}^{in} = \mathop{\arg\max}_{\hat{\mathcal{D}}^{n}:|\hat{\mathcal{D}}^{n}| \leq r |\overline{\mathcal{D}}^{n}|} \sum\nolimits_{x \in \hat{\mathcal{D}}^{n}} P_{in}(x),
	\label{eq5}
\end{equation}
where $ r $ is the relabeling rate. To boost the robustness of the model, we also propose a labeling net $ L_{net} $ which produces pseudo labels $ \hat{y} $ for these selected in-distribution noisy images through:
\begin{equation}
	\hat{y}=L_{net}(f(x);\theta_{l}), x \in \overline{\mathcal{D}}^{in},
	\label{eq6}
\end{equation}
where $ \theta_{l} $ is the parameter of $ L_{net} $. Finally, these relabeled in-distribution noisy images along with clean ones are utilized for training. The parameter $ \theta_{h} $ is updated according to the descent direction of the expected loss as in Eq.~\eqref{eq7}.
\begin{equation}
	\begin{gathered}
	\theta_{h}^{(t+1)} = \theta_{h}^{(t)} - \alpha \nabla(\frac{1}{|\overline{\mathcal{D}}^{c}|+|\overline{\mathcal{D}}^{in}|} (\sum_{x \in \overline{\mathcal{D}}^{c}} \mathcal{L}(x,y;\theta_{h}^{(t)}) +\\ \sum_{x \in \overline{\mathcal{D}}^{in}} \mathcal{L}(x,\hat{y};\theta_{h}^{(t)}))).
	\end{gathered}
	\label{eq7}
\end{equation}
$ S_{net} $ and $ L_{net} $ are trained using the meta set $ \mathcal{D}^{m} $. Their training mechanisms will be detailedly illustrated in the following sections.
	
\subsection{Selection Net $ S_{net} $}
	
The architecture and training mechanism of our selection net $ S_{net} $ are shown in Fig.~\ref{fig1}~(b). To be specific, $ S_{net} $ is an MLP (multilayer perceptron) network with one hidden layer containing 256 nodes. We apply ReLU activation function on each hidden node and utilize Sigmoid activation function to guarantee the output located in the interval of $ [0,1] $. It takes the image features as input and produces the probability that a sample is in-distribution. In our approach, we utilize $ S_{net} $ to distinguish in- and out-of distribution noisy images.
	
We optimize the parameters $ \theta_{s} $ of $ S_{net} $ in a meta-learning based method. Specifically, we first sample a mini-batch $ \overline{\mathcal{D}} $ from the training set, and formulate the classifier learning function $ \hat{\theta_{h}}^{(t)}(\theta_{s}) $ through:
\begin{equation}
	\hat{\theta_{h}}^{(t)}(\theta_{s}) = \theta_{h}^{(t)} - \alpha\frac{1}{|\overline{\mathcal{D}}|}\sum_{x \in \overline{\mathcal{D}}} S_{net}(f(x);\theta_{s}^{(t)}) \nabla_{\theta_{h}} \mathcal{L}(x,y;\theta_{h})|_{\theta_{h}^{(t)}}.
	\label{eq8}
\end{equation}
We utilize the output of $ S_{net} $ as the weight of sample $ x $. In this way, we make the $ \hat{\theta_{h}}^{(t)}(\theta_{s}) $ to be a function of $ \theta_{s} $. Accordingly, we draw a mini-batch $ \overline{\mathcal{D}}^{m} $ from the meta set. It is entered into the classifier net with parameter $ \hat{\theta_{h}}^{(t)}(\theta_{s})  $ to calculate the meta loss. Then, we can update $ \theta_{s} $ via: 
\begin{equation}
	\theta_{s}^{(t+1)} = \theta_{s}^{(t)} - \beta\frac{1}{|\overline{\mathcal{D}}^{m}|}\sum_{x^{m} \in \overline{\mathcal{D}}^{m}} \nabla_{\theta_{s}} \mathcal{L}(x^{m},y^{m};\hat{\theta_{h}}^{(t)}(\theta_{s}))|_{\theta_{s}^{(t)}},
	\label{eq9}
\end{equation}
where $ \beta $ denotes the learning rate of $ S_{net} $. The computation of $ \theta_{s} $ in Eq.~\eqref{eq9} can be obtained by back-propagation with the following derivation:
\begin{equation}
	\begin{aligned}
	&\frac{1}{|\overline{\mathcal{D}}^{m}|}\sum_{x^{m} \in \overline{\mathcal{D}}^{m}} \nabla_{\theta_{s}} \mathcal{L}(x^{m},y^{m};\hat{\theta_{h}}^{(t)}(\theta_{s}))|_{\theta_{s}^{(t)}}
	\\=&\frac{1}{|\overline{\mathcal{D}}^{m}|}\sum_{i=1}^{|\overline{\mathcal{D}}^{m}|}
	\frac{\partial \mathcal{L}(x^{m}_{i},y^{m}_{i};\hat{\theta_{h}}) }{\partial \hat{\theta_{h}}(\theta_{s})}|_{\hat{\theta_{h}}^{(t)}} \ \times
	\\&\ \sum_{j=1}^{|\overline{\mathcal{D}}|}
	\frac{\partial\hat{\theta_{h}}^{(t)}(\theta_{s})}{\partial S_{net}(f(x_{j});\theta_{s})}
	\frac{\partial S_{net}(f(x_{j});\theta_{s})}{\partial\theta_{s}}|_{\theta_{s}^{(t)}}
	\\=&\frac{-\alpha}{|\overline{\mathcal{D}}^{m}|*|\overline{\mathcal{D}}|}
	\sum_{i=1}^{|\overline{\mathcal{D}}^{m}|}
	\frac{\partial \mathcal{L}(x^{m}_{i},y^{m}_{i};\hat{\theta_{h}}) }{\partial \hat{\theta_{h}}(\theta_{s})}|_{\hat{\theta_{h}}^{(t)}} \ \times 
	\\&\ \sum_{j=1}^{|\overline{\mathcal{D}}|}
	\frac{\partial\mathcal{L}(x_{j},y_{j};\theta_{h})}{\partial\theta_{h}}|_{\theta_{h}^{(t)}}
	\frac{\partial S_{net}(f(x_{j});\theta_{s})}{\partial\theta_{s}}|_{\theta_{s}^{(t)}}
	\\=&\frac{-\alpha}{|\overline{\mathcal{D}}|}\sum_{j=1}^{|\overline{\mathcal{D}}|} (\frac{1}{|\overline{\mathcal{D}}^{m}|} \sum_{i=1}^{|\overline{\mathcal{D}}^{m}|} T_{ij}) \frac{\partial S_{net}(f(x_{j});\theta_{s}^{(t)})}{\partial\theta_{s}}|_{\theta_{s}^{(t)}},
	\end{aligned}
	\label{app1}
\end{equation}
where $ T_{ij} = \frac{\partial\mathcal{L}(x_{i}^{m},y_{i}^{m};\hat{\theta_{h}})}{\partial\hat{\theta_{h}}}|^{T}_{\hat{\theta_{h}}^{(t)}} \frac{\partial\mathcal{L}(x_{j},y_{j};\theta_{h})}{\partial\theta_{h}}|_{\theta_{h}^{(t)}} $. Then Eq.~\eqref{eq9} can be rewritten as:
\begin{equation}
	\theta_{s}^{t+1} = \theta_{s}^{(t)} + \frac{\alpha\beta}{|\overline{\mathcal{D}}|}\sum_{j=1}^{|\overline{\mathcal{D}}|} (\frac{1}{|\overline{\mathcal{D}}^{m}|} \sum_{i=1}^{|\overline{\mathcal{D}}^{m}|} T_{ij}) \frac{\partial S_{net}(f(x_{j});\theta_{s}^{(t)})}{\partial\theta_{s}}|_{\theta_{s}^{(t)}}.
	\label{eq10}
\end{equation}
In this formula, the coefficient $ \frac{1}{|\overline{\mathcal{D}}^{m}|} \sum_{i=1}^{|\overline{\mathcal{D}}^{m}|} T_{ij} $ represents the similarity between the gradient of the training sample $ x_{j} $ computed on training loss and the average gradient of the meta data $ \overline{\mathcal{D}}^{m} $ calculated on meta loss. Therefore, if the learning gradient of a training sample is similar to that of the meta images, then it will be considered as in-distribution and $ S_{net} $ tends to produce a higher score for it. On the contrary, samples with gradient different from the that of the meta set will have a lower score. In this way, our $ S_{net} $ learns to leverage the feature representation to distinguish in- and out-of-distribution samples.
	
\subsection{Labeling Net $ L_{net} $}
	
Fig.~\ref{fig1} (c) illustrates our training mechanism of $ L_{net} $. General DNNs are composed of a series of convolutional layers as the feature extractor and a fully connected layer as the classifier. On the basis of this framework, we add an additional fully connected layer as $ L_{net} $ after the last convolutional layer. It takes the image features as input and produces a pseudo label. $ L_{net} $ learns from the clean meta set. Specifically, we first draw a mini-batch $ \overline{\mathcal{D}}^{m} $ from the meta set and utilize the classifier net $ h $ to extract the image features $ f(x^{m}) $ for each sample in $ \overline{\mathcal{D}}^{m} $. Then, we update $ \theta_{l} $ by:
\begin{equation}
	\theta_{l}^{(t+1)}=\theta_{l}^{(t)} - \frac{\alpha}{|\overline{\mathcal{D}}^{m}|} \nabla (\sum_{x^{m} \in \overline{\mathcal{D}}^{m}} \mathcal{L}(L_{net}(f(x^{m});\theta_{l}^{(t)}),y^{m}).
	\label{eq11}
\end{equation}
We adopt the same learning rate for network $ h $ and $ L_{net} $, because $ L_{net} $ and the classifier of $ h $ are parallel fully connected layers (See Fig.~\ref{fig1} (c)), which share the same size of input and output. In this training mechanism, $ L_{net} $ learns to dynamically produce labels through the image features during training.
	
\begin{algorithm}[t]\small
		\SetAlgoLined
		\caption{Data-driven Meta-learning Based Fine-Grained Recognition Algorithm}
		\KwInput{training set $ \mathcal{D} $, meta set $ \mathcal{D}^{m} $, drop rate $\tau$, relabeling rate $r$, batch size $ n $, epoch $T_{s}$, and $ T_{max} $.}
		\textbf{Initialize} classifier network parameter $ \theta_{h} $, $ S_{net} $ parameter $ \theta_{s} $ and $ L_{net} $ parameter $ \theta_{l} $.\\
		\For{$T = 1, 2,..., T_{max}$}
		{
			\For{$ i = 1, 2 , ...,  \frac{|\mathcal{D}|}{n} $}
			{
				Sample a mini-batch $ \overline{\mathcal{D}} $ from $ \mathcal{D} $ and $ \overline{\mathcal{D}}^{m} $ from $ \mathcal{D}^{m} $;\\
				Formulate the learning function $ \hat{\theta_{h}}(\theta_{s}) $ by Eq.~(\ref{eq8});\\
				Update $ \theta_{s} $ by Eq.~(\ref{eq10}) and $ \theta_{l} $ by Eq.~(\ref{eq11});\\
				\eIf{$ T \leq T_{s} $}
				{
					Update $ \theta_{h} $ by Eq.~(\ref{eq1});}
				{
					Obtain $ \overline{\mathcal{D}}^{c} $ via Eq.~(\ref{eq2}) and $ \overline{\mathcal{D}}^{n} $ by Eq.~(\ref{eq3});\\
					Calculate $ P_{in} $ by Eq.~(\ref{eq4}) and $ \overline{\mathcal{D}}^{in} $ via Eq.~(\ref{eq5});\\
					Generate pseudo labels for $ \overline{\mathcal{D}}^{in} $ through Eq.~(\ref{eq6});\\
					Update $ \theta_{h} $ with Eq.~(\ref{eq7});}
			}
		}
		\KwOutput{Updated parameters $ \theta_{h} $, $ \theta_{s} $, and $ \theta_{l} $.}
		\label{algorithm}
\end{algorithm}
	
We directly utilize the feature extractor trained on the noisy dataset and simply leverage a fully connected layer as $ L_{net} $. The aim is to avoid over-fitting and make $ L_{net} $ more generalized. If we pre-train a DNN from the meta set as $ L_{net} $, it will probably suffer from over-fitting, as only a small number of clean images is available \cite{li2017learning}. As a result, it can't produce reliable pseudo labels. On the contrary, in our approach, the feature extractor learns from a large number of web training images. Hence, it is more generalized than that trained from a small dataset. In this way, our $ L_{net} $ doesn't need to extract the image feature by itself and only learns to produce labels from the image features. It is less likely to suffer from over-fitting. Our approach takes full advantage of the meta set by utilizing it to train $ S_{net} $ and $ L_{net} $ for sample selection and loss correction, respectively. We train $ S_{net} $, $ L_{net} $ and classifier net $ h $ simultaneously in an end-to-end manner. With the help of $ S_{net} $ and $ L_{net} $, our approach has the ability to utilize in-distribution noisy images and can re-use clean samples, which may be eliminated in the selection guided by training loss. To conclude, our approach combines the sample selection and loss correction methods and overcomes their drawbacks. It can train the model robustly from noisy web images in the real-world dataset. The detailed steps of our proposed approach are summarized in the Algorithm~\ref{algorithm}.

\begin{table*}[t]\small
	\centering
	\renewcommand{\arraystretch}{1.1}
	\caption{ACA (\%) performances on three benchmark fine-grained datasets. BBox/Anno (\checkmark) indicates human annotations are utilized during training. Training set denotes the dataset is a manually labeled one (anno.), a small clean meta set (meta.), or collected from the web (web).}
	\vspace{-0.3cm}
	\begin{tabular}{c|r|c|c|c|c|c|c}
		\toprule
		\multirow{2}{*}{\textbf{Supervision}} & \multirow{2}{*}{\textbf{Method\:\:\:\:\:\:\:\:\:\:\:\:\:}} & \multirow{2}{*}{\textbf{Publication}} & \multirow{2}{*}{\textbf{BBox/Anno}} & \multirow{2}{*}{\textbf{Training Set}} & \multicolumn{3}{c}{\textbf{Datasets}} \\
		\cline{6-8} &                &                      &          &    & CUB200-2011  & FGVC-Aircraft & Cars-196\\		
		
		\hline
		\multirow{4}{*}{Strongly}
		
		&  Part-Stacked CNN \cite{huang2016part}	& CVPR 2016 & \checkmark  & anno.      &  76.60  &  - & - \\
		&  Coarse-to-fine \cite{yao2016coarse}	& TIP 2016  & \checkmark  & anno.      &  82.90  &  87.70 & - \\
		&  HSnet \cite{lam2017fine}				& CVPR 2017 & \checkmark  & anno.      &  87.50  &  - & 93.90 \\
		&  Mask-CNN \cite{wei2018mask} 		 	& PR 2018   & \checkmark  & anno.      &  85.70  &  - & - \\
		
		\hline
		\multirow{7}{*}{Weakly} 			
		
		&  Bilinear CNN \cite{lin2015bilinear}      & ICCV 2015 &     & anno.      &  84.10  & 83.90 & 91.30\\
		&  RA-CNN \cite{fu2017look}                 & CVPR 2017 &     & anno.      &  85.30  & -     & 92.50\\
		&  Multi-attention \cite{zheng2017learning} & ICCV 2017 &     & anno.      &  86.50  & 89.90 & 92.80\\
		&  Filter-bank \cite{wang2018learning}      & CVPR 2018 &     & anno.      &  86.70  & 92.00 & 93.80\\
		&  Parts Model \cite{ge2019weakly}		   & CVPR 2019 &     & anno.      &  90.40  & -		& -	   \\
		&  TASN \cite{zheng2019looking}	   		   & CVPR 2019 &     & anno.      &  89.10  & -		& 93.80\\
		&  DCL \cite{chen2019destruction}		   & CVPR 2019 &     & anno.      &  87.80  & 93.00	& 94.50\\
		
		\hline
		\multirow{2}{*}{Webly}
		
		&  WSDG \cite{niu2015visual}                  & CVPR 2015    &      & web &  70.61   & -	 & -\\
		&  Xiao \al \cite{xiao2015learning}           & CVPR 2015    &      & web &  70.92   & -	 & -\\
		
		\hline
		\multirow{3}{*}{Semi} 
		
		&  Xu \al \cite{xu2016webly}               & TPAMI 2018 & \checkmark  & anno.+web  & 84.60   & -	 & -\\
		&  Cui \al \cite{cui2016fine}              & CVPR 2016  & \checkmark  & anno.+web  & 80.70   & -	 & -\\
		&  Niu \al \cite{niu2018webly}             & CVPR 2018  &             & anno.+web  & 76.47   & -	 & -\\
		
		\hline
		\multirow{4}{*}{Meta}
		&  Decoupling \cite{malach2017decoupling}     & NeurIPS 2017 &      & meta.+web & 75.04 & 81.88 & 84.78\\
		&  Co-teaching \cite{han2018co}               & NeurIPS 2018 &      & meta.+web & 82.62 & 82.81 & 88.78\\
		&  MW-Net \cite{shu2019meta}				  & NeurIPS 2019 &      & meta.+web & 77.37 & 77.17 & 84.09\\
		& \textbf{Our Approach}                       & -            &      & meta.+web & \textbf{84.19}  &\textbf{83.95} &\textbf{89.83}
		\\
		\bottomrule
	\end{tabular}
	\label{tab1}
	\vspace{-0.1cm}
\end{table*}
 
\section{Experiments}

\subsection{Datasets and Evaluation Metric}

We evaluate our approach on three commonly used benchmark fine-grained datasets, CUB200-2011 \cite{wah2011caltech}, FGVC-Aircraft \cite{aircraft}, and Cars-196 \cite{car196}.\\
Average Classification Accuracy (\textbf{ACA}) is taken as the evaluation metric, which is widely used for evaluating the performance of fine-grained visual classification. 

\subsection{Implementation Details} 

We leverage the web images collected in \cite{AAAI2020} as the noisy training set and directly adopt the testing data from CUB200-2011, FGVC-Aircraft, and Cars-196 as the test set. The small clean meta set is built by randomly sampling 10 images for each category from the benchmark training set CUB200-2011, FGVC-Aircraft, and Cars-196. Accordingly, we have 18388, 13503 and 21448 noisy web training images as well as 2000, 1000 and 1960 small clean meta images for CUB200, FGVC-Aircraft and Cars-196 dataset, respectively. 

We utilize a pre-trained ResNet-34 \cite{he2016deep} model as our backbone network and select the drop rate $ \tau $ from the values of \{0.15, 0.2, 0.25, 0.3, 0.35, 0.4\}. We ultimately set the value of $ \tau $ to be 0.35, 0.20, and 0.25 as the default value on CUB200, FGVC-Aircraft and Cars-196 dataset, respectively. For other parameters, we set relabeling rate $ r=0.05 $ for CUB200 and FGVC-Aircraft dataset, and $ r=0.15 $ for Cars-196 dataset. The initial epoch number $ T_{s} $ is set to be 5 on all three datasets. We use an SGD optimizer with momentum$ =0.9 $, and train our model for 100 epochs with batch size set as 60. The learning rate $ \alpha $ is set to be 0.01, which is decayed with a cosine annealing \cite{loshchilov2016sgdr}, and $ \beta $ is fixed as 0.01.

\subsection{Baseline Methods}

To illustrate the superiority of our approach, our baselines contain the following five sets of state-of-the-art fine-grained methods.\\ 
\textbf{1) Strongly supervised methods:} Part-Stacked CNN \cite{huang2016part}, Coarse-to-fine \cite{yao2016coarse}, HSnet \cite{lam2017fine}, and Mask-CNN \cite{wei2018mask}.\\
\textbf{2) Weakly supervised methods:} Bilinear CNN \cite{lin2015bilinear}, RA-CNN \cite{fu2017look}, Filter-bank \cite{wang2018learning}, Multi-attention \cite{zheng2017learning}, Parts Model \cite{ge2019weakly}, TASN \cite{zheng2019looking}, and DCL \cite{chen2019destruction}.\\
\textbf{3) Webly supervised methods:} WSDG \cite{niu2015visual}, and Xiao \al \cite{xiao2015learning}.\\
\textbf{4) Semi-supervised methods:} Xu \al \cite{xu2016webly}, Niu al \cite{niu2018webly}, and Cui \al \cite{cui2016fine}.\\
\textbf{5) Meta data based methods:} Decoupling \cite{malach2017decoupling}, Co-teaching \cite{han2018co}, and MW-Net \cite{shu2019meta}.\\
It should be noted that all the meta data based methods have the same backbone network ResNet-34 as ours. Their implementations are the same as ours, including learning rate, optimizer and batch size. Moreover, we set the same drop rate as ours for Co-teaching. Experiments are conducted on one NVIDIA V100 GPU card.

\subsection{Experimental Results and Analysis}

Table~\ref{tab1} presents the fine-grained ACA results of various approaches on benchmark datasets. As demonstrated in Table~\ref{tab1}, with the guidance of a small clean meta set, our proposed approach shows significant improvements over web-supervised approaches \cite{niu2015visual,xiao2015learning}. Compared with semi-supervised methods that utilize a large number of manually labeled images \cite{niu2018webly} or even additional human annotations \cite{xu2016webly,cui2016fine}, our approach achieves close or superior performance on CUB200 dataset. Moreover, our approach remarkably surpasses other meta data based methods on all benchmark datasets. 

\begin{table*}[t]
	\begin{minipage}{0.485\linewidth}
		\centering
		\renewcommand{\arraystretch}{1.1}
		\caption{The ACA performances (\%) comparison of different loss correction methods.}
		\vspace{-0.3cm}
		\begin{tabular}{c|c|c} 
			\toprule 
			\textbf{Backbone} & \textbf{Method}  & \textbf{Performance (\%)}   \\
			\hline
			\multirow{5}{*}{ResNet-18}	
			& Discarding (Baseline) 	             & 75.89   \\ 	
			& No-relabeling                          & 74.09   \\ 	
			& Ditillation \cite{li2017learning}      & 74.13   \\
			& Self-correction \cite{song2019selfie}  & 75.41   \\
			& Ours                                   &  \textbf{77.48} \\			
			\bottomrule
		\end{tabular}
		\label{tab2}
	\end{minipage}
	\hspace{0.2cm}
	\begin{minipage}{0.485\linewidth} 
		\centering
		\renewcommand{\arraystretch}{1.1}
		\caption{The ACA (\%) performances comparison of different training set sizes.}
		\vspace{-0.3cm}
		\begin{tabular}{c|c} 
			\toprule 
			\textbf{Number of Images} & \textbf{Performance (\%)}   \\
			\hline
			50 		& 67.35	  \\ 	
			60   	& 71.06   \\ 	
			70      & 72.94   \\
			80      & 75.63   \\
			90      & 77.48   \\			
			\bottomrule
		\end{tabular}
		\label{tab6}
	\end{minipage}
	\vspace{-0.2cm}
\end{table*}

\begin{figure*}[t] 
	\begin{minipage}{0.48\textwidth} 
		\centering 
		\includegraphics[width=1\textwidth]{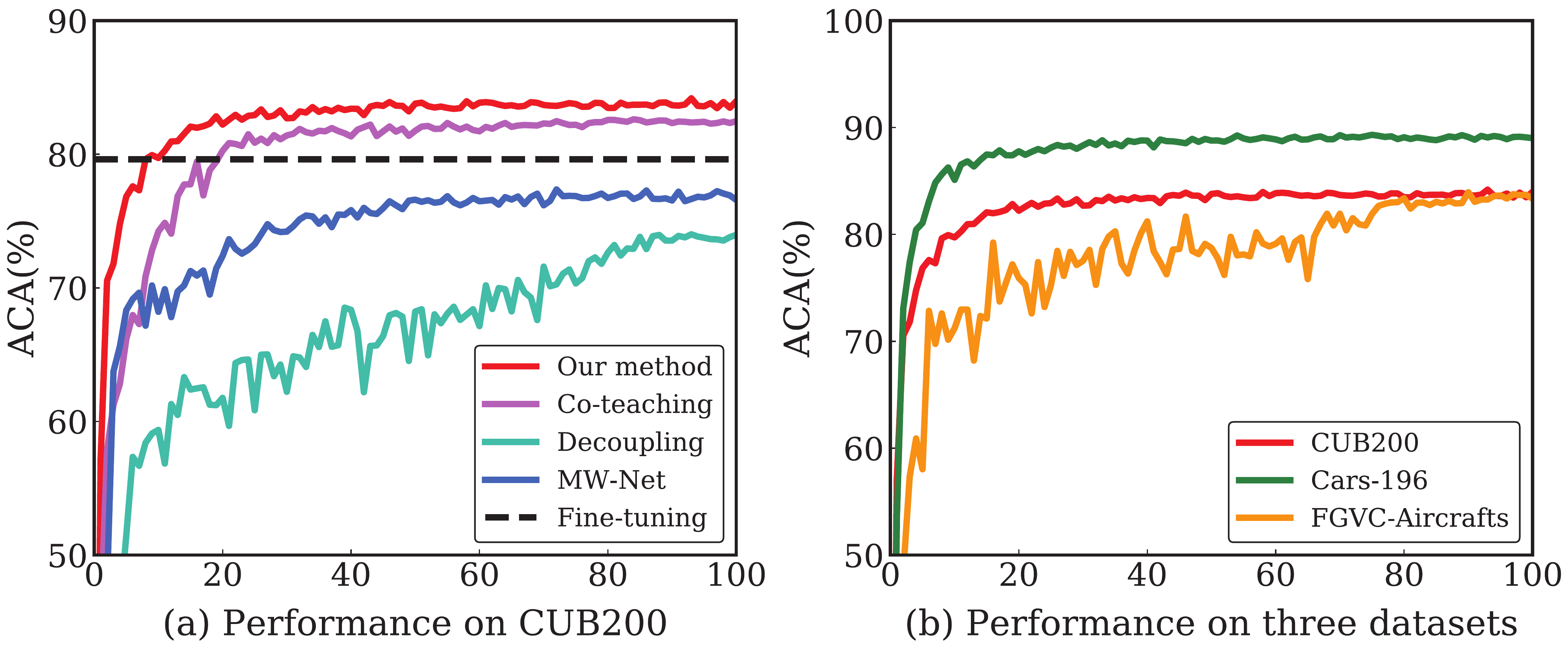} 
		\caption{The test accuracies of baselines and ours on CUB200 dataset (a), and our test accuracies on datasets CUB200, Cars-196 and FGVC-Aircraft dataset (b).}
		\label{fig2}	
	\end{minipage}
	\hspace{0.2cm}
	\begin{minipage}{0.48\textwidth} 
		\centering 
		\includegraphics[width=1\textwidth]{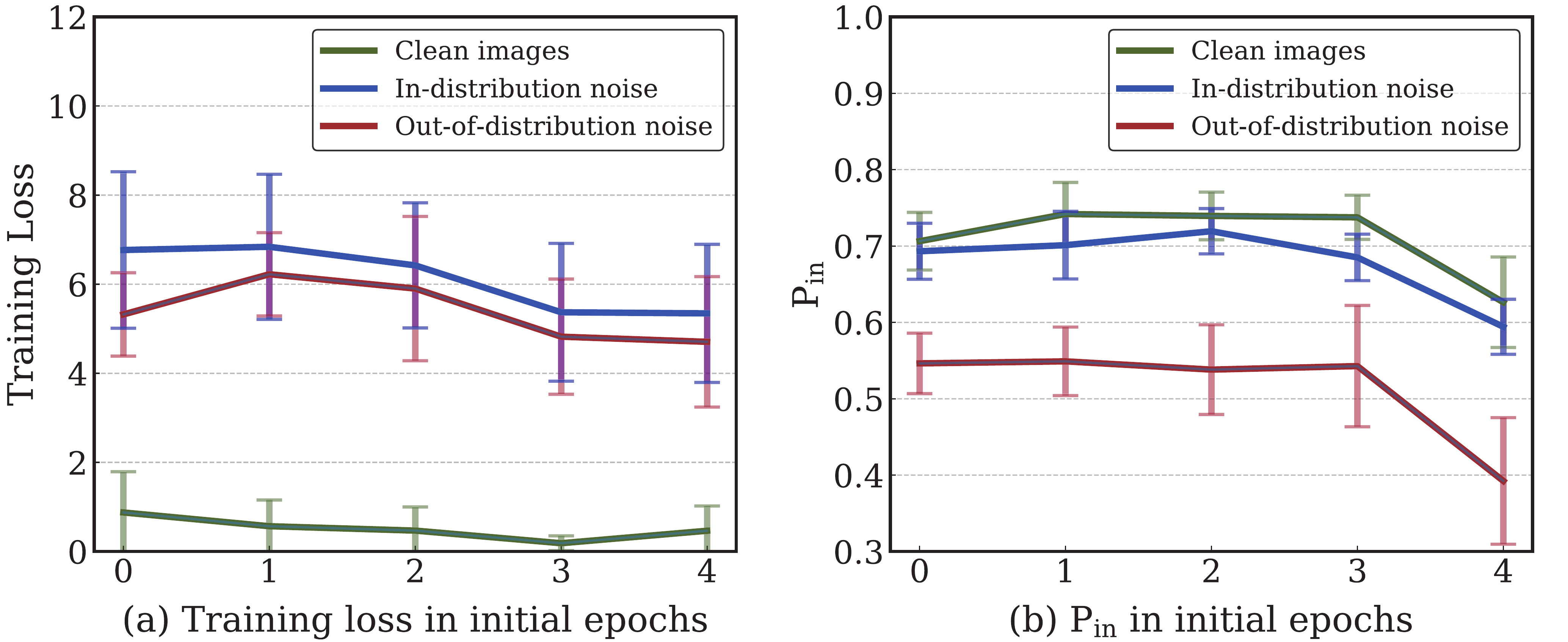} 
		\caption{The training loss (a) and $ P_{in} $ (b) of clean, in-distribution and out-of-distribution noisy images in initial epochs. Bars indicate the standard deviation.}
		\label{fig3}
	\end{minipage}
\end{figure*}

Fig.~\ref{fig2} (a) presents the test accuracies of baselines and ours on CUB200 dataset. By observing Fig.~\ref{fig2} (a), we can notice that our approach shows better performance, training speed and stability than meta data based baselines. MW-Net \cite{shu2019meta} requires a small clean meta set, however, it doesn't take full advantage of the small clean meta set. Compared with MW-Net \cite{shu2019meta}, our approach makes better use of the clean meta set by utilizing it for sample selection and loss correction simultaneously. As a result, our approach outperforms MW-Net significantly on all three benchmark datasets.

From Fig.~\ref{fig2} (a), we can observe that Decoupling \cite{malach2017decoupling} shows the worst performance among all meta data based methods. The reason is that it discards a large number of web training images with which the model can be boosted. We notice that the drop rate of Decoupling climbs as training proceeds, the average value of which is 77\%. In this situation, many clean images are discarded. To overcome this drawback, our approach leverages a fixed drop rate to ensure that most of the images are utilized for training. Co-teaching \cite{han2018co}, which shares the same drop rate as ours, performs best among our baselines except ours. The reason is that our approach can relabel and reuse in-distribution noisy images for boosting the training of the model. 

Fig.~\ref{fig2} (b) presents the test accuracies vs. the number of epochs on all three benchmark datasets. By observing Fig.~\ref{fig2} (b), we can find that the training processes on CUB200 and Cars-196 are fast and stable. This phenomenon demonstrates the superiority of our approach. However, the test accuracy on FGVC-Aircraft fluctuates during training. It may result from that FGVC-Aircraft has the smallest clean meta set among benchmark datasets. A smaller number of trusted images make the selection results and pseudo labels less stable. Although the training process has some fluctuations, the test accuracy becomes stable at the end of training, with a higher value than that of baselines (See Table~\ref{tab1}). From this result, we can conclude that our proposed approach still works even though the clean meta set is very small.

\section{Ablation Studies}

In this section, we will further demonstrate how our approach works. To save time and computing resources, we conduct experiments on the CUB200 dataset with a relatively small pre-trained network ResNet-18 \cite{he2016deep} as our backbone. To better demonstrate the effectiveness of our approach, we remove the small clean meta set from the training set, which means that the meta set is only utilized to train the selection net $ S_{net} $ as well as the labeling net $ L_{net} $, the classifier net $ h $ is only updated by noisy web images.

\begin{table*}[t]
	\centering
	\small
	\begin{minipage}{0.48\linewidth}
		\renewcommand{\arraystretch}{1.1}
		\centering
		\caption{The ACA (\%) performances and improvements of data augmentation. Anno. and web denote the dataset is manually labeled and collected from the web, respectively.}
		\vspace{-0.3cm}
		\begin{tabular}{c|c|c|c} 
			\hline 
			\textbf{Datasets} & \textbf{Training Set} & \textbf{Performance} & \textbf{Improvement}   \\
			\hline
			\multirow{2}{*}{CUB200-2011} 		&     anno.         	& 79.46         & \multirow{2}{*}{$\Delta$ 6.35} \\ 	
			&  	  anno.+web         & \textbf{85.81}         \\
			\hline 
			\multirow{2}{*}{FGVC-Aircraft}  	&     anno.         	& 82.21     & \multirow{2}{*}{$\Delta$ 7.32} \\ 	
			&  	  anno.+web         & \textbf{89.53}         \\
			\hline 
			\multirow{2}{*}{Cars-196} 			&	  anno.				& 88.20		& \multirow{2}{*}{$\Delta$ 3.36} \\	
			&	  anno.+web			& \textbf{91.56}			\\
			\hline
		\end{tabular}
		\label{tab3}
	\end{minipage}
	\hspace{0.5cm}
	\begin{minipage}{0.48\linewidth} 
		\centering
		\renewcommand{\arraystretch}{1.1}
		\caption{The ACA (\%) performances and improvements of ResNet-18, ResNet-34 and ResNet-50. Baseline denotes that the network is directly trained on the web training set.}
		\vspace{-0.3cm}
		\begin{tabular}{c|c|c|c} 
			\hline 
			\textbf{\:\:\:Backbone\:\:\:} & \textbf{\:\:\: Method \:\:\:} & \textbf{Performance} & \textbf{Improvement}   \\
			\hline
			\multirow{2}{*}{ResNet-18}			
			& Baseline &      68.59        	& 
			\multirow{2}{*}{$\Delta$ 8.89} \\ 
			& Ours     &  	  \textbf{77.48}         & \\
			\hline
			\multirow{2}{*}{ResNet-34}	
			& Baseline 	&     74.06        	&  
			\multirow{2}{*}{$\Delta$ 6.51} \\
			& Ours	    &  	  \textbf{80.57}         &  \\
			\hline
			\multirow{2}{*}{ResNet-50}	 
			& Baseline &	  74.92				& 
			\multirow{2}{*}{$\Delta$ 6.84} \\ 
			& Ours     &	\textbf{81.76}  			& \\
			\hline
		\end{tabular}
		\label{tab4}
	\end{minipage}
\end{table*}

\begin{figure*}[t] 
\begin{minipage}{0.48\textwidth} 
\centering
\includegraphics[width=0.63\textwidth]{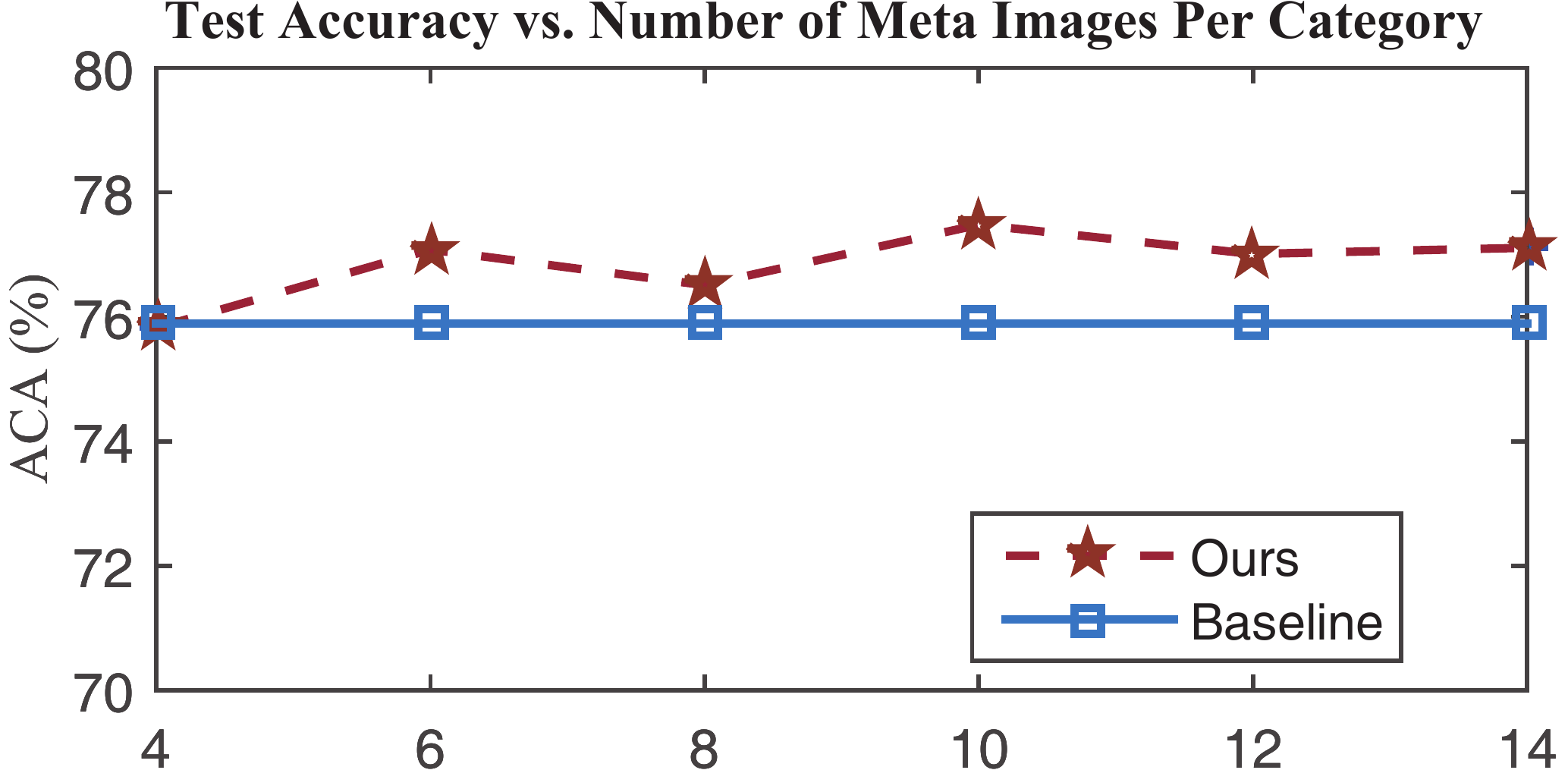}
\caption{The ACA (\%) performances comparison of different meta set sizes.}
\label{fig4}
\end{minipage}
\hspace{0.2cm}
\begin{minipage}{0.48\textwidth} 
\centering 
\includegraphics[width=1\textwidth]{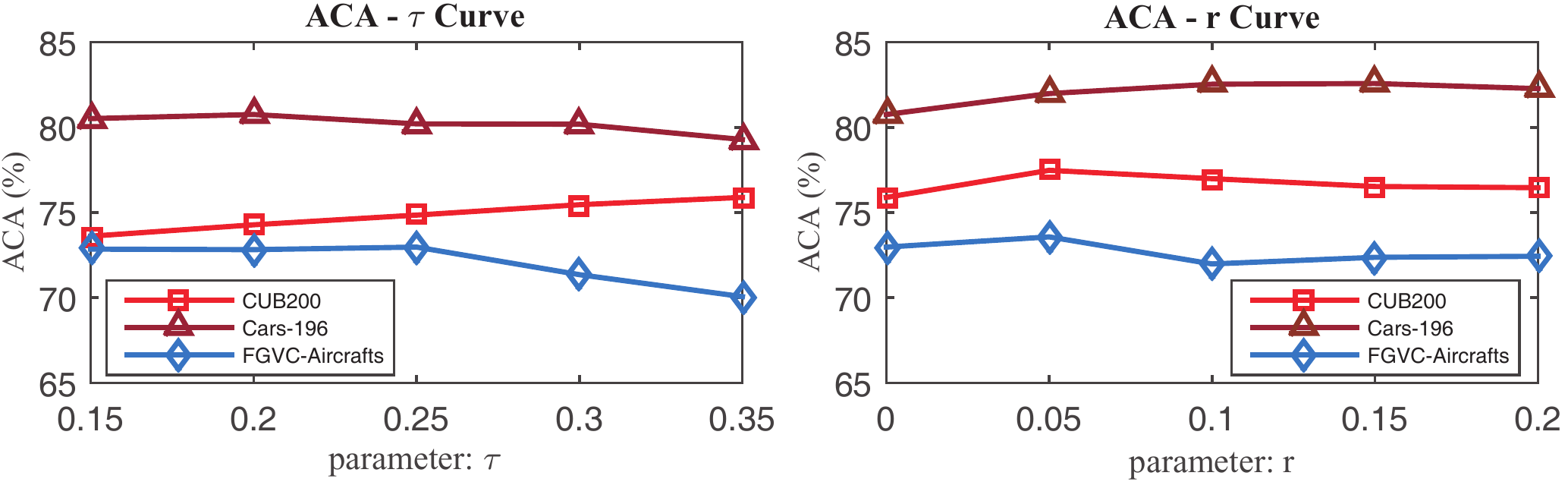} 
\caption{The parameter sensitivities of drop rate $ \tau $ (a) and relabeling rate $ r $ (b).}
\label{fig5}
\end{minipage}
\vspace{-0.3cm}
\end{figure*}

\subsection{Effectiveness of Sample Selection}

In this experiment, we compare the performance of $ P_{in} $ and cross-entropy training loss in the process of samples selection. We first manually add 10 clean images, 10 in-distribution noisy images, and 10 out-of-distribution noisy images into the training set. Then we record their $ P_{in} $ and loss values in the initial epochs during training. The experimental results are shown in Fig.~\ref{fig3}. 

By observing Fig.~\ref{fig3} (a), we can notice that loss values of in- and out-of-distribution noisy images are similar, which is much higher than that of clean images. This phenomenon indicates that utilizing loss values can split clean images from the training set but can't distinguish in- or out-of-distribution noisy samples. From Fig.~\ref{fig3} (b), we can observe that clean and in-distribution noisy images share a close value of $ P_{in} $, while out-of-distribution noisy samples show a lower $ P_{in} $. By leveraging the value of $ P_{in} $, we can distinguish in-distribution ones from noisy images. From Fig.~\ref{fig3} (a) and (b), we can conclude that utilizing loss and $ P_{in} $ simultaneously can efficiently identify clean, in- and out-of-distribution noisy samples. With the ability to distinguish in- and out-of-distribution images, our approach can further utilize in-distribution noisy images by relabeling them.

\subsection{Influence of Different Training Set Sizes}

We investigate the impact of data scale by changing the number of web images per category on CUB200. Specifically, we collect 50, 60, 70, 80, 90 images from the web for each category. As shown in Table \ref{tab6}, the ACA performance improves steadily by using more web training images. Hence, leveraging web images for FGVC is a promising research direction as web data is rich and easy to obtain.

\subsection{Effectiveness of Loss Correction}

To demonstrate the superiority of our labeling net $ L_{net} $, we replace our $ L_{net} $ in our framework with other loss correction approaches for comparison. These loss correction methods are as follows:
1) Discarding (Baseline): Dropping all noisy images and conducting no loss correction;
2) No-relabeling: Utilizing in-distribution noisy images for training without correcting their labels;
3) Distillation \cite{li2017learning}: Pre-training a Resnet-18 model on the validation set for relabeling;
4) Self-correction \cite{song2019selfie}: Utilizing the predictions of classifier network $ h $ as pseudo labels for in-distribution noisy images. In this experiment, we fix the value of $ \tau $ as 0.35 and $ r $ as 0.05.

The experimental results are shown in Table~\ref{tab2}. From Table~\ref{tab2}, we can observe that the model suffers from label noise if the labels of in-distribution noise are not corrected (No-relabeling). Utilizing these noisy images for training is worse than simply discarding them (Baseline). By relabeling in-distribution noisy images, Distillation \cite{li2017learning} and Self-correction \cite{song2019selfie} slightly outperform No-relabeling. Nevertheless, they still show worse performance than the baseline. Their unsatisfying performance results from the correction error. Distillation \cite{li2017learning} is easy to suffer from over-fitting, because the clean meta set only contains 2000 images. The model pre-trained on the small meta set has poor generalization performance and can't produce reliable pseudo labels (Its test accuracy is only 53.33\%). Since fine-grained recognition is challenging, the classifier network tends to give wrong predictions during training, and then Self-correction produces wrong pseudo labels. Our approach overcomes their drawbacks by training $ L_{net} $ with the guidance of the clean meta set. It produces reliable pseudo labels and remarkably outperforms other loss correction approaches.

\begin{figure*}[t]
	\centering 
	\includegraphics[width=0.98\textwidth]{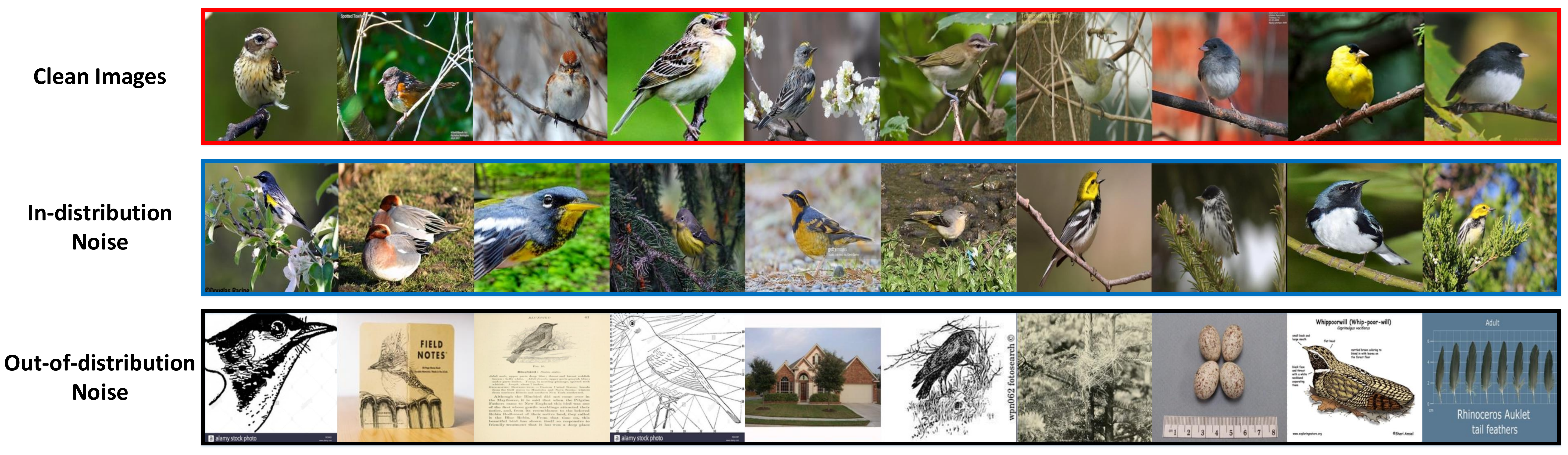} 
	\caption{Sample selection results on noisy web training set in \cite{AAAI2020}. Clean and in-distribution noisy images are similar. Out-of-distribution noisy samples are diverse and totally different from images in the CUB200 dataset.}
	\label{fig6}	
	\vspace{-0.3cm}
\end{figure*}

\subsection{Influence of Data Augmentation}

To demonstrate the advantage of leveraging web images, we train a ResNet-18 model on three benchmark datasets with our web images as data augmentation. Specifically, we utilize the intact manual labeled dataset as the validation set. In addition, we only adopt our algorithm on web images, and manually labeled ones are directly leveraged for training. The results are given in Table~\ref{tab3}. From Table~\ref{tab3}, we can observe that leveraging web images significantly improves the performance across different datasets. The improvements on CUB200, FGVC-Aircraft and Cars-196 are 6.35\%, 7.32\% and 3.23\%, respectively. This result also indicates that leveraging web images is an effective method to improve the robustness of FGVC model.

\subsection{Influence of Different Backbones}

We investigate the applicability of our approach by conducting experiments with different backbone networks on the CUB200 dataset. The experimental results are demonstrated in Table~\ref{tab4}. From Table~\ref{tab4}, it can be observed that our approach shows remarkable improvements across different backbones. Guided by some meta images, networks can learn from web data more efficiently. 

\begin{table}[b]
	\centering
	\renewcommand{\arraystretch}{1.1}
	\caption{The performances of different $ S_{net} $ structures. Baseline indicates discarding all noisy images. FC and numbers denote the fully connected layer, and the number of hidden nodes in the MLP, respectively.}
	\vspace{-0.3cm}
	\begin{tabular}{c|c} 
		\toprule 
		\textbf{$ S_{net} $ Architectures} & \textbf{Performance (\%)}   \\
		\hline
		Baseline 	& 75.89   \\ 	
		FC   		& 76.75   \\ 	
		100         & 76.92   \\
		256         & 77.48   \\
		1024        & 76.58   \\			
		\bottomrule
	\end{tabular}
	\label{tab5}
\end{table}

\subsection{Influence of Different Meta Set Sizes}

In this experiment, we change the number of meta images for each category to investigate the influence of meta set size. The experimental results are illustrated in Fig.~\ref{fig4}. From Fig.~\ref{fig4}, we can find that unless the meta set is extremely small (\eg, each category only has 4 images or less), our approach can outperform the baseline by relabeling in-distribution noisy images. Most importantly, the performance remains roughly stable when the image number increases to 6 or more. This phenomenon indicates that our approach is robust and a small clean meta set can ensure reliability and performance. 

\subsection{Influence of Different $ S_{net} $ Architectures}

In this experiment, we compare the ACA performances of different $ S_{net} $ architectures: a fully connected layer and MLP networks with one hidden layer containing 100, 256 and 1024 nodes. Table \ref{tab5} illustrates the experimental results. By observing Table \ref{tab5}, we can find that all architectures can outperform the baseline and achieve close test accuracies. Even utilizing the simplest fully connected layer as $ S_{net} $ can work well. This result indicates that our approach is robust and doesn't require complex network architectures.

\subsection{Parameter Sensitivities}

For the parameter sensitivities analysis, we study the drop rate $ \tau $ and relabeling rate $ r $ on three datasets. As illustrated in Fig.~\ref{fig5} (a), each dataset has its own optimal drop rate. The best value is 0.35 for CUB200, 0.2 for Cars-196, and 0.25 for FGVC-Aircrafts. Before $ \tau $ increases to the optimal value, the ACA performance rises steadily as $ \tau $ increases. The improvement may benefit from discarding more noisy samples. If $ \tau $ exceeds the optimal value, the performance obviously declines. One possible explanation is that too many instances are discarded and the network can not get sufficient training data. Fig.~\ref{fig5} (b) presents the influence of relabeling rate $ r $. By observing Fig.~\ref{fig5} (b), we can find that the optimal relabeling rate for CUB200, Cars-196 and FGVC-Aircrafts is 0.05, 0.15 and 0.05, respectively. On Cars-196 dataset, the ACA performance climbs as $ r $ increases to the optimal value. It probably results from leveraging more in-distribution noisy images for training. Conversely, the performance drops as $ r $ increases ($ r > 0.05$) on CUB200 and FGVC-Aircrafts dataset. The reason may be that some out-of-distribution samples are utilized and misguide the model as $ r $ rises.

\subsection{Visualization}

To intuitively demonstrate our sample selection ability, we visualize the selection results on noisy web training set \cite{AAAI2020} in Fig.~\ref{fig6}. From Fig.~\ref{fig6}, we can observe that three kinds of samples are clearly separated. In-distribution noisy images are similar to clean ones but are probably mislabeled. Some out-of-distribution samples are related to the bird but totally different from images in manually labeled dataset CUB200, \eg, books or drawings. This selection result demonstrates that our approach is practical and able to deal with the noisy real-world dataset. Moreover, our approach can be utilized to refurbish the web training set by discarding harmful out-of-distribution noisy images.

\section{Conclusion}

In this paper, we presented a data-driven meta-set based method to learn from noisy web images for fine-grained visual classification. Our motivation is to combine the "sample selection" and "loss correction" method, and utilize the in-distribution noisy web images for boosting training. Comprehensive experiments on three real-world scenario datasets demonstrate that our approach is much superior to state-of-the-art meta-set based methods for fine-grained visual classification.

\section*{Acknowledgments}
This work was supported by the National Natural Science Foundation of China (No. 61976116, 61702265, 61932020), National Key R\&D Program of China (No. 2018AAA0102001), Fundamental Research Funds for the Central Universities (No. 30920021135), and Natural Science Foundation of Jiangsu Province (No. BK20170856). 

\bibliographystyle{ACM-Reference-Format}
\bibliography{references}

\end{document}